\documentclass[letterpaper, 10 pt, conference]{ieeeconf}  % Comment this line out if you need a4paper

\IEEEoverridecommandlockouts

\usepackage{flushend}
\usepackage{balance} % Comment 
\let\labelindent\relax

\usepackage{etoolbox} % fix space between abstract header and abstract body
\usepackage[utf8]{inputenc} % allow utf-8 input
\usepackage[T1]{fontenc}    % use 8-bit T1 fonts
\usepackage[english]{babel} %for hyphenation rules
\usepackage[protrusion=true,expansion=true]{microtype}
\usepackage{comment}
\usepackage{cancel}
\usepackage{csquotes}
\usepackage{listings}
\usepackage{nameref}
\usepackage{paralist}
\usepackage{xspace}
\usepackage{setspace}
\usepackage{makeidx}
\usepackage{pdfpages}

\usepackage{amsmath,amssymb} % define this before the line numbering.
\usepackage{amsfonts}       % blackboard math symbols
\usepackage{mathtools}
\usepackage{array} %for table entries to be in center of cell
\usepackage{nicefrac}       % compact symbols for 1/2, etc.
\usepackage{breqn}          % dmath multiline
\usepackage{textcomp}
\usepackage{bm} %bold symbols in math mode
\usepackage{pifont}
\usepackage[
  separate-uncertainty = true,
  multi-part-units = repeat
]{siunitx}

\usepackage{graphicx}
\usepackage{adjustbox} %add additional box operations in addition to resizebox
\usepackage{epsfig}
\usepackage{float}
\usepackage{subcaption}
\usepackage[font=small,labelfont=bf]{caption}
\usepackage{lscape}      %Useful for wide tables or figures.

\usepackage{tikz}  %make graphs
\usepackage{makecell}
\usepackage{overpic}
\usepackage{algorithm}
\usepackage[noend]{algpseudocode}

\usepackage{array} %for table entries to be in center of cell
\usepackage{tabularx}
\usepackage{multirow}
\usepackage{multicol}
\usepackage{booktabs}   % Publication quality tables
\usepackage{tablefootnote}

\usepackage{paralist}
\usepackage{enumitem}
\setlist[itemize]{noitemsep,leftmargin=*,topsep=0in}
\setlist[enumerate]{noitemsep,leftmargin=*,topsep=0in}

\usepackage{url}            % simple URL typesetting
\PassOptionsToPackage{table}{xcolor}
\usepackage{color,colortbl} %superceded by xcolor
\usepackage{soul}

\PassOptionsToPackage{capitalize, noabbrev}{cleveref}
\usepackage[pagebackref=false,breaklinks=true,colorlinks,urlcolor=blue,citecolor=blue,linkcolor=blue,bookmarks=false]{hyperref}

\makeatletter
\let\NAT@parse\undefined
\makeatother
\usepackage[numbers,sort&compress]{natbib}

\usepackage{blindtext}
\usepackage{bbm}

\usepackage{lipsum}

\usepackage{titlesec}
\titlespacing{\section}{0pt}{0.3\baselineskip}{0.25\baselineskip}
\titlespacing{\subsection}{0pt}{0.2\baselineskip}{0.15\baselineskip}
\titlespacing{\subsubsection}{0pt}{0.05\baselineskip}{0.03\baselineskip}

\renewcommand{\paragraph}[1]{\vspace{0.2em}\noindent\textit{#1} --}
\usepackage{listings}
\definecolor{codegreen}{rgb}{0,0.6,0}
\definecolor{codegray}{rgb}{0.4,0.4,0.4}
\definecolor{codepurple}{rgb}{0.5,0,0.9}
\definecolor{backcolour}{rgb}{0.95,0.95,0.95}

\definecolor{lightgray}{rgb}{0.9,0.9,0.9}
\definecolor{lightpink}{rgb}{0.98,0.85,0.86}
\definecolor{lightblue}{rgb}{0.68,0.84,0.9}

\lstdefinestyle{mystyle}{
    backgroundcolor=\color{backcolour},   
    commentstyle=\color{codegreen},
    keywordstyle=\color{magenta},
    numberstyle=\tiny\color{codegray},
    stringstyle=\color{codepurple},
    basicstyle=\fontsize{7.5}{8}\selectfont\ttfamily\ttfamily,
    breakatwhitespace=false,         
    breaklines=true,
    breakindent=0pt,
    captionpos=b,                    
    keepspaces=true,                 
    numbers=none,                    
    numbersep=5pt,                  
    showspaces=false,                
    showstringspaces=false,
    showtabs=false,                  
    tabsize=2
}
\newcommand{\etal}{\textit{et al}.}

\definecolor{color1}{rgb}{.6,.4,.05}
\definecolor{color2}{rgb}{0,.7,.7}
\definecolor{color3}{rgb}{0.35,0.75,0.0}
\definecolor{color4}{rgb}{0.4,0.8,0}
\definecolor{color5}{rgb}{1,0,0}

\usepackage{xcolor}
\usepackage{fancyvrb}
\usepackage{mdframed}
\usepackage{environ}

\DefineVerbatimEnvironment{grayverbatim}{Verbatim}{
  fontsize=\footnotesize,
  frame=single,
  framesep=0.5mm,
  rulecolor=\color{gray!10},
  fillcolor=\color{gray!10},
  framerule=0pt
}

\NewEnviron{grayminipage}{%
  \par\noindent
  \colorbox{gray!10}{%
    \begin{minipage}{\linewidth}
      \ttfamily\footnotesize\BODY
    \end{minipage}%
  }%
  \par
}

\newcommand{\project}{CLIMB\xspace}
\newcommand{\website}{\href{https://plan-with-climb.github.io/}{https://plan-with-climb.github.io/}\xspace}

\newcommand{\ourdomain}{BlocksWorld++}

\title{\LARGE \bf \project:
Language-Guided Continual Learning for \\ Task Planning with Iterative Model Building
}

\author{Walker Byrnes$^{1,2}$, Miroslav Bogdanovic$^{4}$, Avi Balakirsky$^{3}$, Stephen Balakirsky$^{2}$, Animesh Garg$^{1,4,5}$% <-this % stops a space
\thanks{$^{1}$Georgia Institute of Technology, $^{2}$Georgia Tech Research Institute,$^{3}$Ohio State University, $^{4}$University of Toronto, $^{5}$Nvidia}%
\thanks{Correspondence to: \href{mailto:animesh.garg@gatech.edu}{animesh.garg@gatech.edu}}%
}

\begin{document}

\maketitle
\thispagestyle{empty}
\pagestyle{empty}

\begin{abstract}

Intelligent and reliable task planning is a core capability for generalized robotics, requiring a descriptive domain representation that sufficiently models all object and state information for the scene.
We present \project, a continual learning framework for robot task planning that leverages foundation models and execution feedback to guide domain model construction. \project can build a model from a natural language description, learn non-obvious predicates while solving tasks, and store that information for future problems.
We demonstrate the ability of \project to improve performance in common planning environments compared to baseline methods. We also develop \textit{\ourdomain} domain, a simulated environment with an easily usable real counterpart, together with a curriculum of tasks with progressing difficulty for evaluating continual learning. Code and additional details for this system can be found at \website.

\end{abstract}

\section{Introduction}

For decades roboticists have pursued the aim of generalized, flexible robots that can solve complex tasks in novel environments. Recent advances in foundation models \cite{hu2023generalpurposerobotsfoundationmodels} have shown promising results for their use in world modeling \cite{liu2023_reflect, wong2023learningadaptiveplanningrepresentations, han2024interpretinteractivepredicatelearning}, task planning \cite{wang2024describeexplainplanselect, wang2023voyageropenendedembodiedagent, chen2024lasp, silver2024}, and motion planning \cite{liang2023codepolicieslanguagemodel, ding2023tampwithllms, chen2024autotamp}. These works leverage the extensive background knowledge present in the foundation model's pre-training to provide incomplete but useful solutions to these challenging tasks. Though incomplete, results can be further refined through repetition, prompt engineering, or post-processing to improve success rates.

While interest in fundamental research has been plentiful, foundation model-guided planners have yet to find regular use in practical application scenarios. This hesitance has been largely derived from the inconsistencies of output for many foundation models. Some research has shown that the direct application of foundational models for task planning yields subpar results \cite{kambhampati2024llmscantplan}, where critics argue foundation models provide an ``approximate retrieval'' of information and are incapable of explicit logical reasoning or planning. This observation suggests that a more sophisticated structure is required in order to leverage the extensive corpus of foundation models effectively.

In this paper, we present \project, Continual Learning for Iterative Model Building. \project is a hybrid neuro-symbolic planning system that makes use of both foundation models and classical symbolic planners to achieve planning proficiency. Given the limitations of both classical search-based planners and end-to-end learning models, a neuro-symbolic approach is required to address complex planning problems reliably. Such a hybrid architecture  leverages both the structure of symbolic planning and the learning capabilities and extensive knowledge base of foundation models effectively. Additionally, \project incrementally builds a PDDL model of its operating environment while completing tasks, creating a set of world state predicates that function as a representation of the causal structure present in the environment. This continual learning approach enables \project to solve types of problems it has previously encountered without the need to relearn task-specific information and endows it with the ability to expand its environment representation to novel problem formulations. We show \project to be a moderately capable planner independently, but importantly we demonstrate its ability to self-improve, resulting in superior performance once a PDDL model has been established.

\begin{figure}[!t]
    \includegraphics[width=\columnwidth]{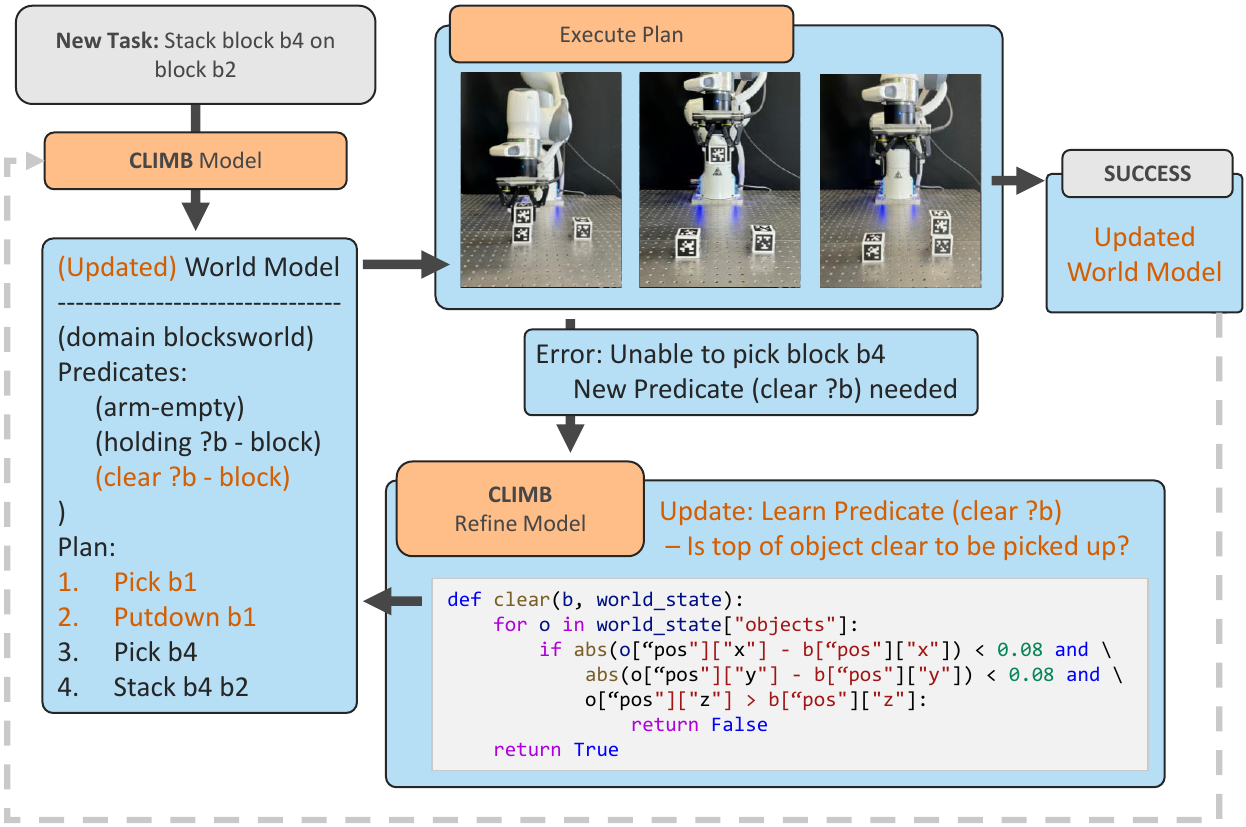}
    \caption{\project is able to predict common world constraints without ground truth PDDL and can learn domain-specific relationships through the execution of one or more tasks in the domain. The framework can self-improve by storing domain and predicate information across tasks.}
    \label{fig:money}
\end{figure}

The contributions of our work are as follows:
\begin{enumerate}
    \item We propose \project for learning logical models of the world and accompanying grounding functions, starting from simple domain and task descriptions in natural language and learning and improving through interaction with the environment.
    \item We evaluate \project to generate sensible initial domain proposals, to improve  through interaction and iterative world model building across several tasks, while proposing and correcting grounding functions to connect the logical domain to the continuous environments.
    \item We propose \ourdomain with a curriculum of tasks to evaluate incremental logical world model-building capabilities, both in simulation and in the real world.
\end{enumerate}

\section{Related Work}

\subsection{Learning-based Planning in Robotics}

Learning methods have long been used to model and represent the object and state interactions in robot environments. Such pursuits have utilized a range of methods including unsupervised clustering \cite{konidaris2018skills, james2022autonomous}, supervised learning from demonstration \cite{silver2023predicate}, and neural architectures \cite{ahmetoglu2022deepsym}. The level of human setup and interaction significantly varies between approaches, with both heavily supervised interaction-based approaches \cite{han2024interpretinteractivepredicatelearning} and fully-unsupervised systems \cite{nguyen2021self} seeing use.

In addition to learning world relationships and predicates, the task of grounding continuous state observations to logical predicates remains a challenge. Both images \cite{mao2019neurosymbolicconceptlearnerinterpreting} and synthesized state representations \cite{Xu_2017_CVPR} have been used in evaluating world state representations. While both approaches have shown success, the former requires large quantities of annotated data while the latter necessitates extensive engineered systems that do not generalize easily.

\subsection{General Planning Capabilities of LLMs}

Recent advances in Large Language Models (LLMs) have shown promising results in their ability to complete logical reasoning tasks, decompose complex goals into sub-goals, and ingest large amounts of data. These capabilities are aided by research into prompting strategies and in-context learning \cite{wei2022chain, yao2023treeofthoughts, Besta_2024graphofthoughts, yao2023reactsynergizingreasoningacting}, evaluating the best methods for interacting with these models. Chain-of-Thought \cite{wei2022chain}, tree-of-thoughts \cite{yao2023treeofthoughts}, graph-of-thoughts \cite{Besta_2024graphofthoughts}, and ReAct \cite{yao2023reactsynergizingreasoningacting} examine the relative performance of prompting structures for complex reasoning tasks. While LLMs can effectively generate reasonable unstructured natural language outputs, they have demonstrated poor performance at structured and constrained tasks including planning \cite{kambhampati2024llmscantplan} without additional structure. These limitations also occur in embodied environments. SayCan \cite{ahn2022saycan} and Reflexion \cite{shinn2023reflexion} propose action embedding and verbal RL to surprising success on text-based tasks. Unstructured natural language tasks do not easily translate into most embodied robot paradigms which frequently utilize discrete actions or task policies with structured interpretations.

Additionally, significant attention has been given to LLMs applicability towards embodied reasoning in grounded environments \cite{ahn2022saycan, singh2022progpromptgeneratingsituatedrobot}. Recent research demonstrates some ability for LLMs to solve some classes of planning problems directly. However, planning performance of these models is inconsistent at best and insufficient at worst, with state of the art planning architectures still critically failing in some domains (e.g. the \textit{floortile} domain in \cite{liu2023llmp}). This has led to significant debate about whether LLMs are capable of planning in a traditional sense \cite{kambhampati2024llmscantplan}. Hybrid approaches which utilize both foundation models and symbolic planners are gaining in popularity, as they are able to leverage the strengths of both approaches \cite{guan2023leveragingpretrainedlargelanguage, liu2023llmp, han2024interpretinteractivepredicatelearning, Ding_2023, chen2024lasp}.

\subsection{Continual/Lifelong Learning for Planning}

Lifelong or Continual Learning (CL) presents a significant challenge for training wherein all data is not collected a priori \cite{lesort2019continuallearningroboticsdefinition}. It is essential for robots operating in complex and loosely structured environments (e.g. in the home) to augment their training sets as they operate. Mendez \etal \cite{mendezmendez2023embodiedlifelonglearningtask} have looked at CL paradigms utilizing diffusion models as samplers in a bi-level TAMP framework. They propose the CL problem as one of compositionality \cite{mendez2022reuse}, as opposed to most other research which follows more traditional train and test datasets.

\begin{figure*}[!t]
    \includegraphics[width=\textwidth]{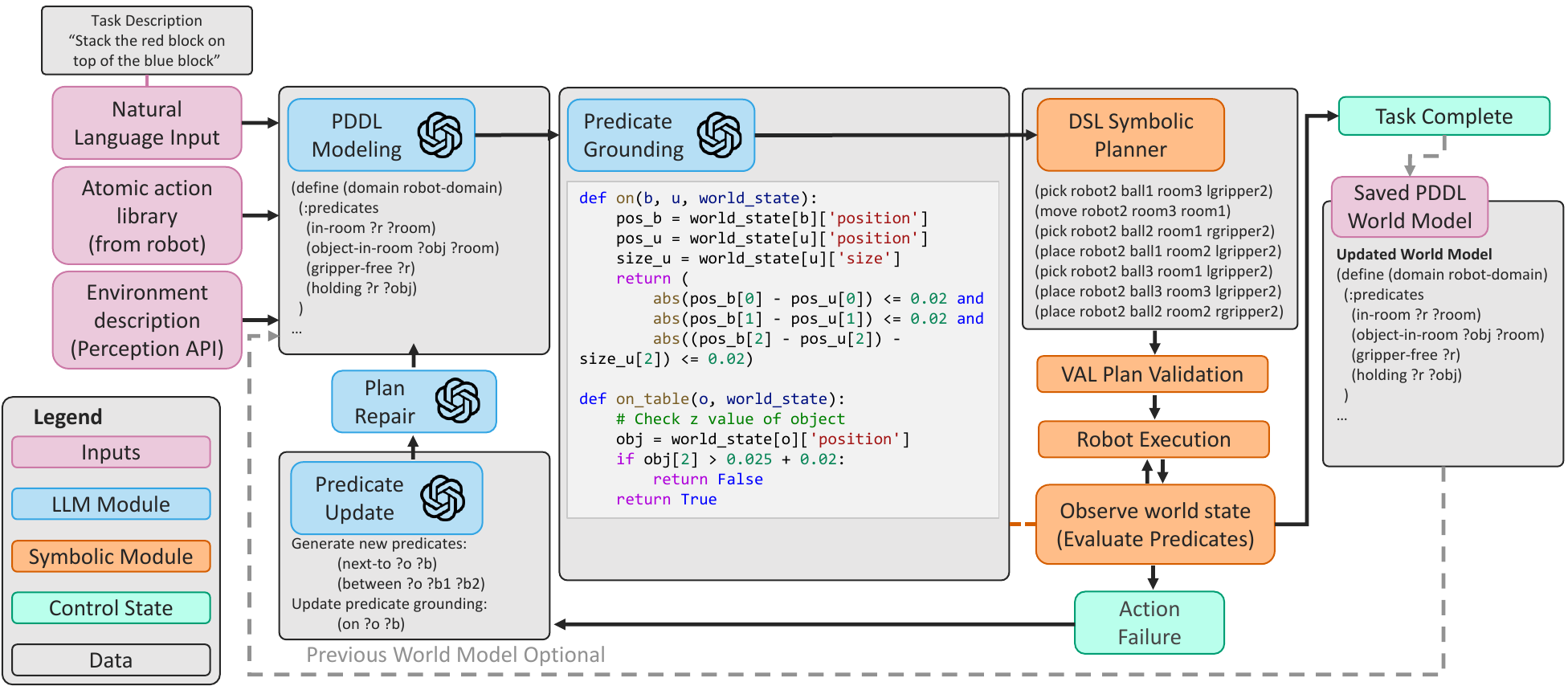}
    \caption{\textbf{The CLIMB Planning Framework} includes multiple independent modules for problem translation, planning, predicate generation, verification, execution, and perception. }
    \label{fig:architecture}
\end{figure*}

\section{Problem Statement}

\noindent\project framework utilizes the following as input:
\begin{enumerate}
    \item \textbf{Domain description:} Given in natural language, in order to provide the general context of the environment the system is operating in.
    \item \textbf{Logical actions:} A set of predefined low-level primitives that the system can use to interact with the world, defined as python function signatures.
    \item \textbf{Tasks to complete:} Given in natural language, a set of tasks for the system to complete and represent into its single logical world model. These tasks also serve as the curriculum to enables the logical planner to find a generalized representation that can solve new instances of such tasks. 
    \item \textbf{(optional) Previous world model:} In case the approach has already been run in the same domain, we can take the resulting logical world model and expand it with additional predicates for completing more varied and complex tasks.
\end{enumerate}

\noindent Using these inputs, \project achieves two goals:
\begin{enumerate}
    \item \textbf{Solving the planning tasks:} Through repeated execution in the domain, analysis of failures, and fixing errors in generated plans or predicates, the goal of the system is to incrementally solve the entire list of tasks while acquiring knowledge about how the environment functions through the process.
    \item \textbf{Building an incremental logical world model:} While solving individual tasks the system reuses and expands upon the logical world model produced from previous tasks. Upon completion of all individual tasks, we are left with a model that can be used to planning solutions for new instances of similar tasks. This model (domain and predicates) can be used as a prior in future executions of this pipeline on new sets of tasks with the same robot embodiment. 
\end{enumerate}

\section{\project: Model Structure}

\paragraph{\textbf{Overview}} The overall architecture for the framework is presented in Figure \ref{fig:architecture}. \project is comprised of modules that generate the PDDL, construct a plan trace for the given problem, observe the robot's performance, and refine the PDDL through observation and queried solutions from the LLM. Each of the LLM modules utilizes the \texttt{gpt-4o-2024-08-06} model from OpenAI. 

Implementation details and full prompts for each of the modules can be found at \website.

\subsection{Language-Model-Guided Domain Generation}

\noindent \project uses four modules to interface with an LLM:

\paragraph{\textbf{Domain and Problem Generation}}
The domain-specific language (DSL) generation module converts a given domain and problem description in unstructured natural language to structured PDDL representations. This generated DSL functions as an estimated representation of a zero-shot world model for the task planner. The initial generation process draws inferences about the logical predicates governing the environment based exclusively on the general world knowledge contained in the LLM and the user's description of the domain. 

\begin{grayminipage}
\textbf{Input: }
\begin{enumerate}
    \item Domain description (natural language)
    \item Atomic actions (function signatures and code)
    \item (optional) Stored domain from previous tasks
\end{enumerate}

\textbf{Output:}
\begin{enumerate}
    \item Domain specific PDDL
    \item List of domain predicates to implement
\end{enumerate}
\end{grayminipage}

\paragraph{\textbf{Predicate Grounding and Debugging}}
This module is used to generate executable Python functions that convert the continuous world state observed from the perception API into a logical predicate set. Grounding predicates allows us to compare the state of the world after each attempted action to the simulated logical state represented by the PDDL model. The comparison of simulated to perceived logical state serves as the primary error identification mechanism by which previously unseen relationships and constraints can be modeled by the predicate inventor. This module also automatically debugs and corrects syntax errors in the predicate grounding functions through analysis of error messages and performing function regeneration if needed.

\begin{grayminipage}
\textbf{Input: }
\begin{enumerate}
    \item Domain description (natural language)
    \item Continuous state representation (python dict)
    \item Existing predicate names and functions
    \item Names and descriptions of new predicates
\end{enumerate}

\textbf{Output:}
\begin{enumerate}
    \item New predicate names and evaluation functions
\end{enumerate}
\end{grayminipage}

\begin{figure*}[!t]
    \includegraphics[width=\textwidth]{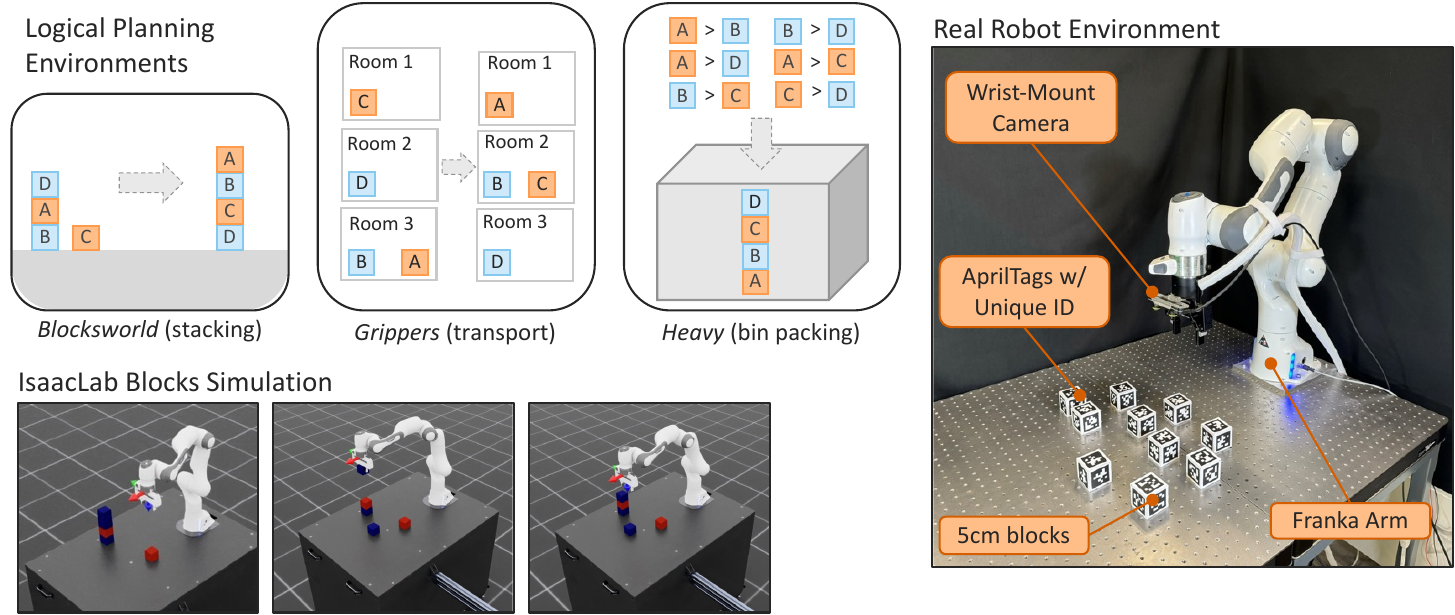}
    \caption{We evaluate \project across logical, simulated, and real domains. Logical domains (BlocksWorld, Grippers, and Heavy) provide a convenient mechanism for comparing performance across a variety of robot embodiments. Experiments in IsaacLab and Real environments extend the block manipulation setting to include more complex stacking and arranging tasks. The simulated and real also serve to evaluate predicate grounding and learning from real experience.}
    \label{fig:experiments}
\end{figure*}

\paragraph{\textbf{Predicate Update}}
When an unexpected logical state is observed after executing an action, the Predicate Update module generates plausible explanations for the phenomena in the form of new or modified predicates. This module's inputs include the current predicate set, the problem and plan that is currently being investigated, and the world state in which the error occurred. From this information, the language model is instructed to reason through potential root causes and proposes an updated predicate set to represent dynamics in the domain. Any new or modified predicates are then re-grounded through the Predicate Grounding module.

\begin{grayminipage}
\textbf{Input:}
\begin{enumerate}
    \item Generated PDDL domain and problem
    \item Generated PDDL plan
    \item World state (before and after execution)
    \item Natural language error message
\end{enumerate}

\textbf{Output:}
\begin{enumerate}
    \item New predicate names and descriptions to resolve error
\end{enumerate}
\end{grayminipage}

\paragraph{\textbf{Plan Repair}}
After an execution failure and predicate generation, the Plan Repair module updates the domain representation to include new predicates and constraints. Inputs to this module are the previous domain and problem DSL, the natural language task description, the world state at failure, and the updated predicate set. This module outputs an updated DSL domain which includes any new or modified predicates and an updated problem file to reflect the current world state. By updating the problem DSL, the system can recover from unintended consequences of executing infeasible actions. For example, if a stack of blocks is knocked over during execution, the initial problem state will be modified to reflect the fact that these blocks are no longer stacked.

\begin{grayminipage}
\textbf{Input: }
\begin{enumerate}
\item Task description (natural language)
\item Generated domain and problem PDDL
\item List of available actions
\item List of perceived objects in the world
\item Names and descriptions of new predicates to add
\end{enumerate}
\textbf{Output:}
\begin{enumerate}
    \item Improved PDDL domain and problem utilizing new predicates
\end{enumerate}
\end{grayminipage}

\subsection{Planning \& Perception}

\paragraph{\textbf{PDDL Symbolic planner}}
To generate the plan trace for a given problem, we make use of the FastDownward symbolic planning framework \cite{Helmert2006_fastdownward} in this study. FastDownward (FD) takes as input the domain and problem PDDL files generated for a specific problem and outputs a plan trace for execution. Internally FD implements a best-first search algorithm with a ``causal graph heuristic'' which is derived from the recursive decomposition of subtasks towards the goal. It is worth noting that our architecture can make use of any symbolic PDDL planner; it has additionally been tested using the Fast-Forward \cite{hoffmann2001ff} and Pyperplan \cite{alkhazraji-et-al-zenodo2020} planning systems. After a plan has been calculated, we additionally verify syntactic correctness and adherence to the generated PDDL by processing the output on the VAL plan validator \cite{Howey2004VALAP}. VAL is a logical plan simulator which utilizes the domain and problem file given to evaluate if the plan is valid within the domain and correctly solves the task. While VAL is not able to check the semantic correctness of the domain and problem, it is capable of ensuring the plan's validity for the given domain. We use the Unified Planning library \cite{unified-planning} for its implementations of all of the above planners, VAL integration, and simulated plan rollout.

\begin{grayminipage}
\textbf{\ourdomain~Action Set:}
\begin{itemize}
    \item pickup(block o)
    \item putdown(block o)
    \item stack(block o, block under)
    \item unstack(block o, block under)
    \item place_between(block o, block a, block b)
    \item stack_on_two(block o, block a, block b)
    \item place_in_front_of(block o, block ref)
    \item place_behind(block o, block ref)
    \item place_to_right_of(block o, block ref)
    \item place_to_left_of(block o, block ref)
\end{itemize}
\end{grayminipage}

\paragraph{\textbf{Plan Execution and Skill Library}}
The plan execution module wraps a library of executable atomic skills and enables the execution of generated plans on the environment. The library of skills is specific to the robot's morphology and to an extent, its domain of operation. However, the skill library should ideally generalize to a variety of tasks within the same environment. By way of example, above are the actions utilized in our \textit{\ourdomain} experimental domain.

\paragraph{\textbf{Perception API}}
While the robot executes the plan trace generated to solve the tasked problem, a domain-specific perception API is utilized to evaluate if the robot accomplished each action successfully. For the real robot system, we utilize AprilTag2 \cite{wang2016apriltag} to observe the position of objects in the environment and distinguish between blocks. This perception module can be substituted for any system that tracks and outputs the 6D pose of objects (e.g. FoundationPose \cite{wen2024foundationposeunified6dpose} for unlabeled objects). After each action is attempted, the perception module collects the state of all objects in the workspace and the proprioception of the robot. This continuous world state representation is then evaluated on the generated predicate functions created by the Predicate Inventor to determine the estimated logical world state.

\section{Experiments}

The goal of our experiments aims to evaluate the following hypothesis: Can LLMs create robust DSL domains and problems without feedback from human experts? How effective are LLMs at representing common world predicates? Can feedback for predicate learning be generated automatically and interpreted by the LLM without human intervention? Do the environment models generated by \project generalize to new problems within the domain?

We evaluate these questions across \textit{logical}, \textit{simulated}, and \textit{real} domains, each with separate characteristics.

\begin{table}[!t]
\centering
\small
\setlength{\tabcolsep}{4pt}
\caption{\textbf{Performance on the Logical Planning Domains} \project demonstrates an ability to generate correct zero-shot plans in some cases, but importantly the system is able to improve its representation over successive (maximum of five) attempts to significantly improve performance ($N=60$ for each case).}
\label{table:logical-results}
\begin{tabular}{lccc}
\toprule
\rowcolor[HTML]{CBCEFB} 
\textbf{Dataset} & LLM Plan & \project 0-Shot & \project Few-Shot \\
\midrule
\texttt{BLW} & 0.12 & 0.40 & \textbf{0.80} \\
                     & \footnotesize{(0.05, 0.20)} & \footnotesize{(0.28, 0.53)} & \textbf{\footnotesize{(0.70, 0.90)}}  \\
\rowcolor[HTML]{EFEFEF} 
\texttt{GRP} & 0.10 & 0.53 & \textbf{0.93} \\
\rowcolor[HTML]{EFEFEF} 
                 & \footnotesize{(0.03, 0.18)} & \footnotesize{(0.42, 0.67)} & \textbf{\footnotesize{(0.87, 0.98)}} \\
\texttt{HVY} & \textbf{0.68} & 0.17 & \textbf{0.67} \\
               & \textbf{\footnotesize{(0.57, 0.80)}} & \footnotesize{(0.08, 0.27)} & \textbf{\footnotesize{(0.55, 0.78)}} \\
\bottomrule
\end{tabular}
\end{table}

\subsection{Logical Planning Domains}

We first evaluate the performance of our architecture on three logical plan-level domains: \textit{blocksworld} \cite{liu2023llmp}, \textit{grippers} \cite{liu2023llmp}, and \textit{heavy} \cite{silver2024}. These domains provide reasonable scenarios for generating manipulator plans and provide points of comparison to related work. As these domains are already logical (i.e. state of the world our system has access to is logical), here we are not evaluating the predicate grounding capabilities of our system. Instead, it allows us to evaluate the capabilities of the system to propose the initial domain, improve it based on failure in executing a plan, and incrementally improve the domain across several tasks in the domain.

\paragraph{\textbf{Domain Description}}

\begin{itemize}
    \item \texttt{BlocksWorld} This domain tasks the agent to stack and unstack columns of blocks to match specific configurations. A popular foundational planning problem, it is likely some examples of this type of environment are present in the LLM training corpus.
    \item \texttt{Grippers} This domain tasks multiple robots to pick up and transport objects between several different rooms or areas. This domain includes a logical representation for robot position requiring the plan to coordinate multiple robots with one another.
    \item \texttt{Heavy} This domain tasks the agent to pack a box or crate with objects, sorting by weight so that heavier objects are placed below lighter objects.
\end{itemize}

\paragraph{\textbf{Correctness of DSL built with Iterative Model Building}}
Experiments in the logical domain serve to evaluate the ability of LLMs to create world representations in DSL. We evaluate both the zero-shot and few-shot performance of \project incorporating up to five rollouts (i.e. executions) and iterations of feedback correction. In the logical domain, we substitute the execution and perception loop with the VAL plan validator \cite{Howey2004VALAP} using a ground truth PDDL domain and problem. VAL models and simulates plan execution in the logical predicate space and evaluates both if the overall plan was successful and, if unsuccessful, where a given error occurs. 

\begin{table}[!t]
\begin{center}
\caption{\textbf{BlocksWorld Predicate Grounding Performance} The LLM demonstrates a capability to generate correct grounding functions for common predicates in BlocksWorld. Functions were evaluated both for \texttt{syntactic} correctness and \texttt{semantic} correctness with respect to examples with known ground truth. We find that the relative frequency of syntax errors is generally low, but we are able to debug to improve performance.}
\label{table:predicate_success}
\resizebox{\linewidth}{!}{%
\begin{tabularx}{\linewidth}{Xcc}
  \toprule
  \rowcolor[HTML]{CBCEFB} 
  Predicate & Zero-Shot & With Syntax Fixing \\
  \midrule
 \texttt{on-table} & 0.95 & \textbf{1.00} \\
  \rowcolor[HTML]{EFEFEF} 
 \texttt{on} & 0.25 &\textbf{ 0.45} \\
 \texttt{holding} & 0.50 & \textbf{0.65} \\
  \rowcolor[HTML]{EFEFEF}
 \texttt{arm-empty} & 0.10  & \textbf{0.35} \\
 \texttt{clear} & 0.60 & \textbf{0.80} \\ 
  \bottomrule
\end{tabularx}
}%resizebox
\end{center}
\end{table}

Table~\ref{table:logical-results} presents the overall success rate in each of the three evaluated domains along with 95\% confidence intervals. Each domain was evaluated over three iterations of 20 problems, in both zero-shot (maximum one rollout) and few-shot (maximum five rollouts) scenarios. The results in Table~\ref{table:logical-results} demonstrate a moderately capable planner in a zero-shot setting with an overall performance of 37\%. However, overall performance increases to 80\% when the \project framework can be used to repair and improve the domain from failures.

\paragraph{\textbf{Effect of Continual Learning on Performance}}

\begin{figure}[!t]
    \includegraphics[width=\columnwidth]{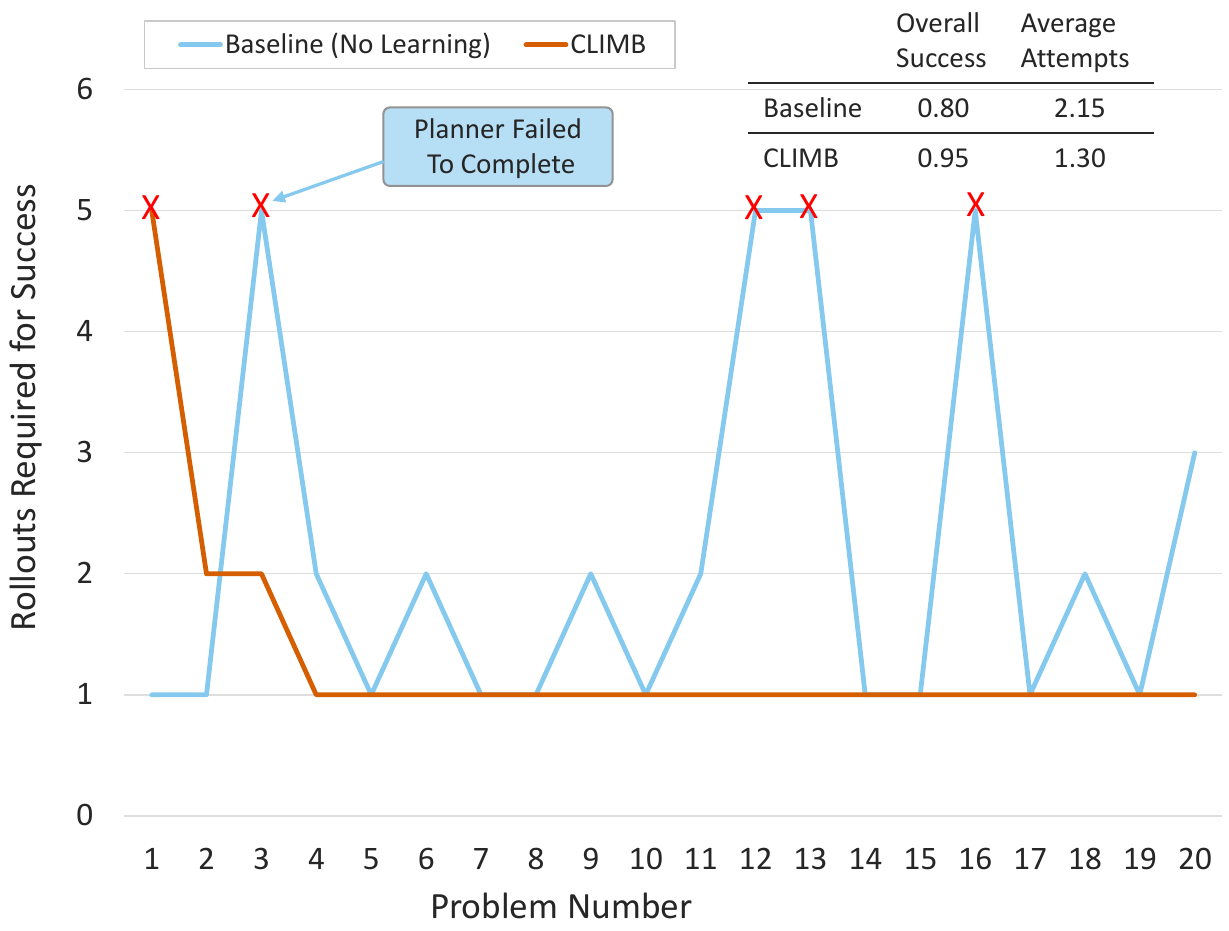}
    \caption{\textbf{Evaluation of the continual learning} capabilities of the \project framework on the \textit{BlocksWorld} logical dataset. The baseline in this case is the same planner without saving domain and predicate information between problems. \project with continual learning significantly outperforms the baseline.}
    \label{fig:continuous-learning}
\end{figure}

\begin{figure*}[!t]
    \includegraphics[width=\textwidth]{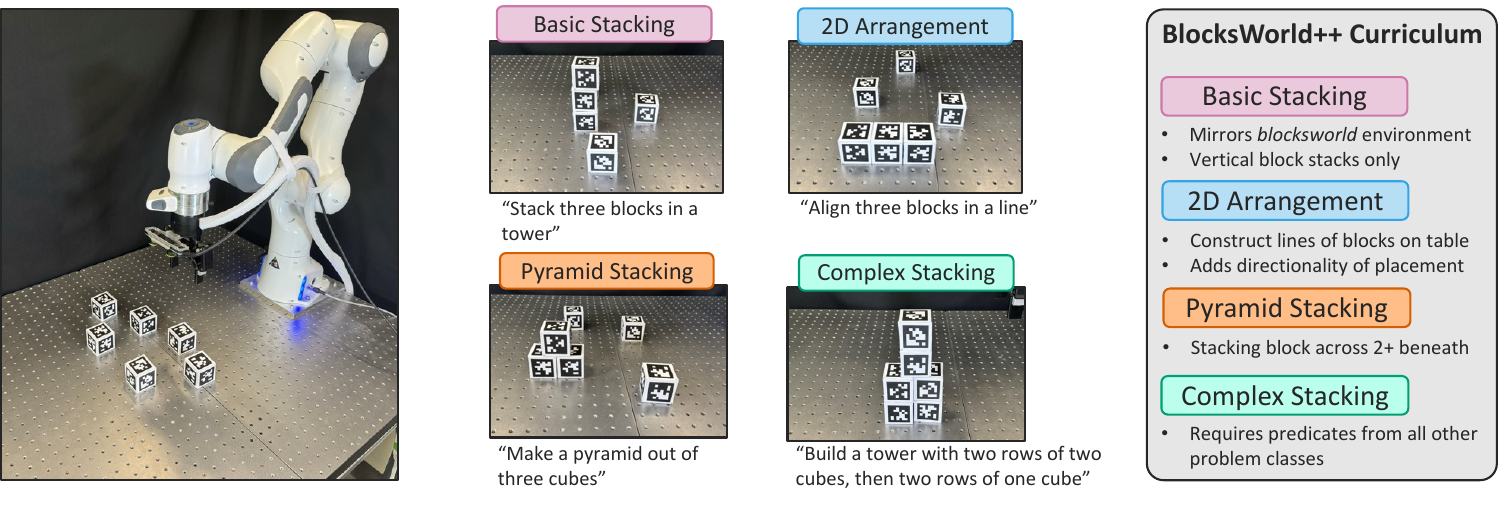}
    \caption{Expanding on the BlocksWorld problem formulation, we have developed the \textit{\ourdomain} dataset which adds additional levels of complexity to the traditional BlocksWorld environment. It includes relative 2D placement on the table and stacking across multiple blocks.}
    \label{fig:real}
\end{figure*}

\begin{figure}[!t]
    \includegraphics[width=\columnwidth]{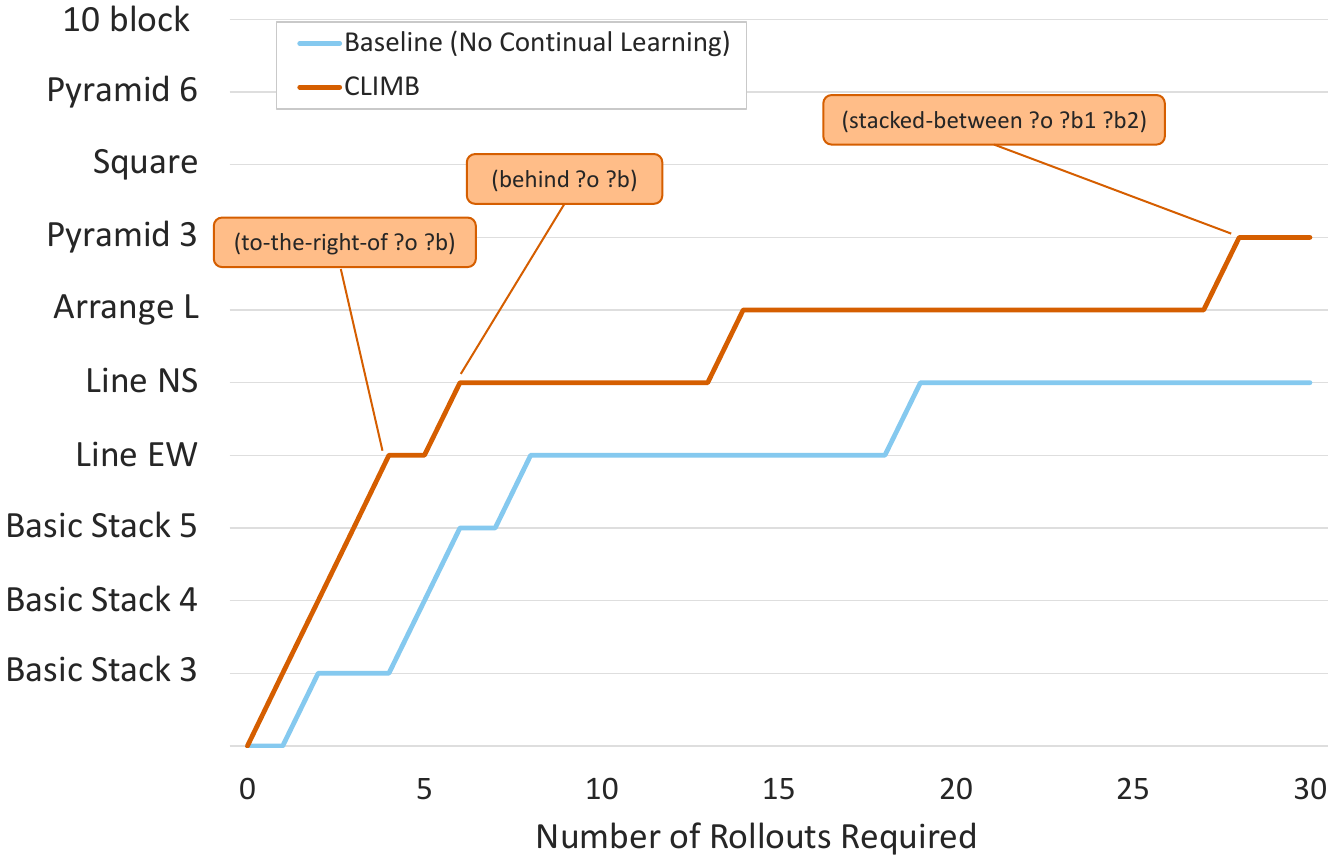}
    \caption{Evaluation of \project's ability to generalize to related tasks in the \textit{\ourdomain} dataset. We label enabling predicates which allow the planner to succeed on new categories of problem. Over 40 iterations, CLIMB is successfully able to solve 7 of the 10 \textit{\ourdomain} problems.}
    \label{fig:domain-generalization}
\end{figure}

To evaluate the ability of \project to produce generalized domain representations, we conduct a comparison of \project's continual learning with a baseline. In the baseline case the planner is still able to execute and learn predicates as in the full pipeline, however results are not saved between tasks. The results of this generalization experiment is shown in Figure~\ref{fig:continuous-learning}. By caching the learned domain and predicates between runs, \project is able to achieve better performance and use 40\% less rollouts to accomplish all tasks in the dataset. 

\subsection{Simulated Robot World}

\paragraph{\textbf{\ourdomain: IsaacLab Block Manipulation Domain}} We have developed an implementation and extension of the BlocksWorld problem in Nvidia IsaacLab~\cite{mittal2023orbit}. This environment enables us to evaluate our approach on the BlocksWorld-type domain, but with continuous state variables and more complex tasks requiring new predicates. We can evaluate whether a correct logical state can be extracted from ground truth continuous state information, and whether this mapping can be iteratively improved based on feedback from plan execution in the domain.

\paragraph{\textbf{Predicate Grounding Function Generation}} To evaluate our predicate generator, we construct grounded predicates from the correct ground truth PDDL domain. We then initialize the IsaacLab environment, extract the scene information, and evaluate our generated predicates to construct a perceived logical world state. This perceived world state is compared to the ground truth PDDL problem initial state for accuracy. We evaluate predicate grounding in isolation by asking it to generate grounding functions for a list of predicates from a given domain and then comparing the values to the ground truth logical values along a single task execution. Table~\ref{table:predicate_success} we show results for 20 independent predicate set generations. Automated syntax error fixing provides only a minor benefit in performance. The majority of errors are caused by semantic, rather than syntactic issues.

\paragraph{\textbf{Domain Generalization on BlocksWorld++ Dataset}} Figure~\ref{fig:domain-generalization} demonstrates the ability for \project to expand it's predicate set to include knowledge relevant to different classes of problems with the same robot embodiment. Evaluating on the \textit{BlocksWorld++} dataset, \project is able to quickly learn simple 1D arrangement in lines on the table. Though more iterations are required, the system is also able to learn spanning across two blocks for the 3 block pyramid problem and more complex horizontal arrangements including an L shape (combination of line NS and line EW). 

\subsection{Physical Robot in the Real-World}

Finally, we demonstrate the system in the real world using the \textit{\ourdomain} domain. Figure \ref{fig:real} showcases the \textit{\ourdomain} curriculum for continual learning evaluation on real hardware. This environment can serve as a benchmark to evaluate \project along with other continual learning paradigms. Moving from simulation to real, the additional challenges are perception and stochastic motion control. To address the challenge of perception, we integrate AprilTag2 \cite{wang2016apriltag} markers on the cubes and a camera mounted on the robot gripper. By returning to a home position after each action, we can gather full pose information for all objects in the scene. We show examples using this system to evaluate \textit{\ourdomain} tasks at \website.

\section{Conclusion}

In this paper we presented \project, a system for incremental learning of logical domains and continuous-state grounding functions through interaction. It requires only a general description of the domain and tasks that need to be solved and access to a set of low-level primitives. \project generates an initial planning domain and logical representation of the task, uses a symbolic planner to solve it, and then learns new world constraints through observation of discrepancies in the logical state expected and observed through the predicate grounding functions.

We evaluate the method capabilities across several logical domains, showing excellent performance given the limited information given as a prior. That, through iterative world interaction and refinement, \project can learn non-intuitive predicates and world constraints to improve performance across successive attempts. Furthermore, we develop the \textit{\ourdomain} dataset, enabling evaluation of continual learning frameworks in a unified manner across logical, simulated, and real domains. 

\clearpage
\renewcommand*{\bibfont}{\small}
\bibliographystyle{IEEEtran.bst} 
\bibliography{climb}

\end{document}